\documentclass{article}
\usepackage[main,final]{neurips_2026}

\usepackage[utf8]{inputenc}
\usepackage[T1]{fontenc}
\usepackage{hyperref}
\usepackage{url}
\usepackage{booktabs}
\usepackage{amsfonts}
\usepackage{nicefrac}
\usepackage{microtype}
\usepackage{graphicx}
\usepackage{xcolor}
\usepackage{xspace}
\usepackage{multirow}
\usepackage{colortbl}
\usepackage{amsmath}
\usepackage{amssymb}
\usepackage{mathtools}
\usepackage{inconsolata}      
\usepackage{microtype}
\usepackage{soul}
\usepackage{subcaption}
\usepackage{graphicx}
\usepackage{enumitem}
\usepackage{arydshln}
\usepackage{bm}
\usepackage{makecell}
\usepackage{array}
\usepackage{wrapfig} 
\usepackage{arydshln}
\usepackage{pgfplots}
\usepgfplotslibrary{groupplots}
\pgfplotsset{compat=1.18}
\usepackage{array}
\usepackage{pifont}
\usepackage{listings}
\usepackage{tabularx}
\usepackage{fontawesome5}
\usepackage{hyperref}

\newcommand{\module}[1]{%
  \rowcolor{gray!13}
  \multicolumn{3}{@{}l}{\textbf{\textit{#1}}} \\
}
\newcommand{\subcfg}[1]{\hspace{0.8em}#1}

\definecolor{categorygray}{gray}{0.93}

\title{Gaussian Splatting-based Volumetric Video Compression with Sparse 4D Anchors}

\author{
\makebox[\textwidth][c]{%
\textbf{Ge~Gao}$^{1}$,
\textbf{Siyue~Teng}$^{1}$,
\textbf{Changqi~Wang}$^{1}$,
\textbf{Fan~Zhang}$^{1}$} \\
\makebox[\textwidth][c]{%
\textbf{Nantheera~Anantrasirichai}$^{1}$,
\textbf{Jui~Chiu~Chiang}$^{2}$,
\textbf{Wen-Hsiao~Peng}$^{3}$,
\textbf{David~Bull}$^{1}$} \\
$^{1}$Visual Information Lab, University of Bristol, UK \\
$^{2}$National Chung Cheng University, Taiwan \space $^{3}$National Yang Ming Chiao Tung University, Taiwan \\
\small \texttt{\{ge1.gao, siyue.teng, changqi.wang, fan.zhang, n.anantrasirichai, dave.bull\}@bristol.ac.uk} \\
\small \texttt{rachel@ccu.edu.tw, wpeng@cs.nycu.edu.tw} \\[3pt]
{\small
\href{https://ge1gao.github.io/saga/}{\textcolor{lilac}{\faGlobe\ \textbf{Project Page}}}
\quad
\href{https://github.com/ge1gao/SAGA-NeurIPS26}{\textcolor{lilac}{\faGithub\ \textbf{Code}}}
}
}

\newcommand{\name}{SAGA\xspace}
\newcommand{\fst}{\cellcolor{red!20}}
\newcommand{\snd}{\cellcolor{orange!20}}
\newcommand{\trd}{\cellcolor{yellow!25}}

\definecolor{lightblue}{HTML}{01feff}
\definecolor{darkerblue}{HTML}{807fff}
\definecolor{brightpurple}{HTML}{ff00ff}
\definecolor{cmarkcolor}{HTML}{168b23}
\definecolor{xmarkcolor}{HTML}{b2100e}
\definecolor{cvprblue}{HTML}{118ab2}
\definecolor{teal}{HTML}{047c84}
\definecolor{tiger}{HTML}{fa6a03}
\definecolor{lilac}{HTML}{dc317f}
\definecolor{bristol}{HTML}{ab1f2e}
\definecolor{metablue}{HTML}{0264e0}
\definecolor{lightgreen}{HTML}{EBFFEB}
\definecolor{lightred}{HTML}{FFEBEB}

\definecolor{memOld}{RGB}{55,105,160}     
\definecolor{memWrite}{RGB}{190,105,45}   
\definecolor{memGate}{RGB}{135,85,155}    
\definecolor{memNew}{RGB}{45,140,85}      

\newcommand{\OldMem}[1]{\textcolor{memOld}{#1}}
\newcommand{\WriteMem}[1]{\textcolor{memWrite}{#1}}
\newcommand{\GateMem}[1]{\textcolor{memGate}{#1}}
\newcommand{\NewMem}[1]{\textcolor{memNew}{#1}}

\newcommand{\cmark}{\textcolor{cmarkcolor}{\ding{51}}}%
\newcommand{\xmark}{\textcolor{xmarkcolor}{\ding{55}}}%

\begin{document}

\maketitle

\begin{figure}[h]
    \centering
    \includegraphics[width=\textwidth]{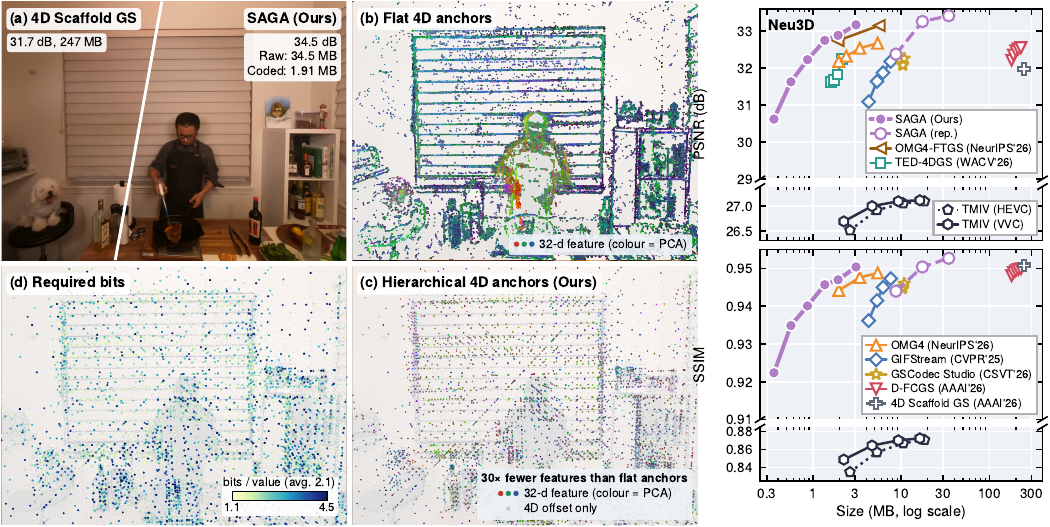}
     \caption{\textbf{SAGA: Hierarchical 4D anchors for compact dynamic scenes.} While flat 4D anchors (\textbf{b}) store features individually, SAGA (\textbf{c}) concentrates features at sparse coarse-level anchors, achieving a $30\times$ reduction in feature-bearing anchors. Our long-context entropy model further exploits dependencies among features, achieving an average coding rate of just 2 bits/symbol (\textbf{d}). Together, these designs compress a 300-frame Neu3D scene to 1.91MB---$129\times$ smaller than 4D Scaffold-GS (\textbf{a})---while achieving the best rate-distortion trade-off among state-of-the-art baselines (\textbf{right}).}
\end{figure}

\begin{abstract}
Immersive video communication requires photorealistic, compact dynamic scene representations that support efficient rendering. 3D Gaussian Splatting (3DGS) is a promising approach, but dynamic 3DGS remains difficult to compress due to its dense primitives and spatiotemporal redundancy. Anchor-based formulations improve compactness through sparse scaffolds that share geometry and appearance across primitives. However, existing designs typically deform a single canonical scaffold and condition each primitive solely on its associated anchor, limiting their ability to model non-local dynamics and disocclusions and exploit inter-anchor correlations. To address these limitations, we propose \textcolor{bristol}{\textbf{\name}}, a volumetric video codec built on \textcolor{bristol}{\textbf{S}}parse \textcolor{bristol}{\textbf{A}}nchor-assisted \textcolor{bristol}{\textbf{GA}}ussian splatting representations. \name represents a dynamic scene using hierarchical sparse anchors in 4D space-time. Coordinate-based INR decoders generate fine anchors and Gaussian primitives from features obtained through kernel interpolation among nearby anchors, enabling parameter sharing across spatiotemporal structures. A fixed-size memory with a novel orthogonality-informed slot-update strategy further captures long-range dependencies among unstructured anchors for entropy coding. Empirically, \name achieves PSNR-based BD-rate reductions of \textbf{90.88\%}, \textbf{64.30\%} and \textbf{65.14\%} relative to GIFStream, TED-4DGS, and OMG4, respectively, on the Neu3D testset.
\end{abstract}

\section{Introduction}
Immersive visual content is increasingly important for applications such as telepresence, free-viewpoint video, virtual and augmented reality, and digital media production \cite{allie2025comprehensive,boyce2021mpeg}. It provides a stronger sense of presence, richer spatial understanding, and greater viewpoint flexibility than conventional 2D media. However, these benefits introduce substantial challenges. While 2D video mainly records view-specific projected appearance on the imaging plane, free-viewpoint volumetric video requires dynamic scene representations that support rendering from multiple viewpoints, further increasing the demands on storage, transmission, and rendering. Consequently, compact dynamic scene representation has become central to scalable immersive video systems, where fidelity, rendering efficiency, and compressibility must be jointly optimized.

For compact representation and compression of volumetric visual content, prior work spans traditional geometry- and video-based pipelines~\cite{li2024mpeg,li2024mpeg-gpcc,allie2025comprehensive,boyce2021mpeg} and neural scene representations~\cite{kwan2024mvhinerv,zhu2025implicit,ling2025multi}. Among neural methods, Neural Radiance Fields (NeRF) have advanced novel-view synthesis~\cite{mildenhall2021nerf} and compression~\cite{wu2024tetrirf,zheng2024hpc,hu2025vrvvc,zhang2024rateaware} by modeling scenes as continuous volumetric radiance fields, but their reliance on volume rendering remains costly for real-time immersive applications. In contrast, 3D Gaussian Splatting (3DGS)~\cite{kerbl20233dgs} uses an explicit, rasterization-friendly representation with efficient training, high visual fidelity, and substantially faster rendering. Consequently, 3DGS compression (e.g., \cite{lu2024scaffold,hac2024}) has become an active research direction for reducing storage costs and enabling practical immersive visual communication.

Following advances in static 3DGS compression \cite{lu2024scaffold,hac2024,chen2025hac++,chen2025fast,wang2024contextgs,navaneet2024compgs}, GS-based volumetric video coding has attracted increasing attention \cite{sun20243dgstream,hu20254dgc,li2025gifstream}. Compared with static scenes, dynamic 3DGS must represent changes in geometry, appearance, and visibility over time, introducing additional parameters and spatiotemporal redundancy. Existing methods broadly fall into frame-by-frame approaches \cite{sun20243dgstream,hu20254dgc,lee2025compression,zhang2026d}, which encode temporally correlated Gaussian frames, and canonical-to-deformation approaches \cite{chen_4dgs-cc_2025,li2025gifstream,ho2026ted} that model video- or chunk-wise temporal variations from a shared canonical representation. While effective, frame-wise methods exploit temporal redundancy via predictive or correlated coding, which can complicate random access and accumulate drift under long prediction chains; canonical-to-deformation methods rely on a single reference that can be affected by abrupt motion, disocclusion, or topology changes.

To address these limitations, we propose \textbf{\name}, a Gaussian Splatting-based volumetric video compression framework using a hierarchical sparse 4D anchor scaffold. \name organizes anchors directly in spacetime, allowing features to be shared across spatial and temporal neighborhoods without deforming a single canonical scaffold. Fine anchors and Gaussian primitives are generated through a unified \emph{interpolation-and-decoding} mechanism: neighboring parent-anchor features are aggregated at each target coordinate and fed, together with positional embeddings, into shared coordinate-based implicit neural representation (INR) decoders. This enables parameter sharing at multiple granularities while retaining local flexibility to represent dynamic scene content. A lightweight local hash-based residual branch further refines primitive attributes in normalized anchor coordinates to recover high-frequency details. To complement this local feature sharing, we introduce a long-context entropy model that summarizes previously decoded anchor information in a fixed-size memory bank. Orthogonality-guided slot updates incorporate new context while limiting repeated writes of information aligned with the previous slot state. Our main contributions are as follows:

\begin{itemize}[leftmargin=0pt, itemindent=1em]
    \item We propose a novel \textbf{GS-based volumetric video compression framework} with hierarchical sparse 4D anchors. By organizing anchors directly in spacetime, our formulation combines local adaptability to geometry and visibility changes with long-range temporal sharing, enabling bidirectional redundancy reduction and flexible non-canonical modeling of complex dynamics.
    
    \item We leverage \textbf{coarse-to-fine and inter-anchor correlations} through position-interpolated hierarchical features. These features condition both fine-anchor and Gaussian decoding, allowing them to aggregate neighboring-anchor context rather than relying on isolated anchor features.
    
    \item We introduce an efficient \textbf{long-context entropy model} for unstructured 4D anchors. It distills decoded spatiotemporal contexts into a compact memory bank whose slots are updated with an orthogonality-guided rule, capturing long-range dependencies under bounded complexity.
\end{itemize}

Experiments on Neu3D \cite{li2022neural}, Panoptic Sports \cite{joo2015panoptic}, and MPEG MIV Common Test Condition (CTC) \cite{boyce2021mpeg} datasets show that \name achieves favorable rate-distortion performance against existing GS-based and MPEG-standard volumetric video codecs. 

\section{Related Work}

\noindent \textbf{Volumetric video compression} has traditionally focused on explicit scene data and multi-view signals, such as point clouds and multi-view videos. Representative standardization efforts include V-PCC \cite{schwarz2019emerging,jang2019video}, G-PCC \cite{li2024mpeg-gpcc}, and V3C \cite{allie2025comprehensive}, covering video-based and geometry-based coding of volumetric content. For immersive multi-view video, TMIV \cite{allie2025comprehensive,boyce2021mpeg} provides reference software that exploits inter-view redundancy and packs texture and geometry into video atlases. Video-based volumetric codecs can further leverage mature 2D coding tools and ecosystems, such as HEVC \cite{sullivan2012overview} and VVC \cite{bross2021overview}, with performance typically assessed under common test conditions and view-synthesis evaluation protocols. Beyond standardized pipelines, learning-based methods have also been explored for explicit 3D data, especially point clouds, using learned geometry transforms, entropy models, temporal prediction, motion compensation, and residual coding \cite{quach2019learning,fan2022ddpcc,jiang2023end,sheng2026dcvc}. More recently, coordinate-based and implicit neural representation (INR) codecs have emerged for multi-view and immersive video compression, such as MV-HiNeRV \cite{kwan2024mvhinerv}, MV-IERV \cite{zhu2025implicit}, and MV-MGINR \cite{ling2025multi}.

\noindent \textbf{Compression for NeRF-based methods} has evolved along a clear methodological progression. Early work improved compactness through end-to-end optimization \cite{bird20213d} and more structured radiance-field parameterizations, including sparse grids, explicit voxels \cite{zhao2023tinynerf}, tensor decompositions \cite{tang2022compressible,chen2022tensorf}, multiresolution hash encodings \cite{muller2022instant}, and plane-based factorization \cite{cao2023hexplane,fridovich2023k}. Later works treat NeRFs more directly as compressible representations, introducing transform coding \cite{rho2023masked,lee2024ecrf}, quantization \cite{li2023compressing,kang2025codecnerf}, entropy-constrained coding \cite{lee2024ecrf,li2024nerfcodec}, and learned codec regimes \cite{li2024nerfcodec,kang2025codecnerf,wang2023neural,fang2024acrf,zheng2024hpc,hu2025vrvvc}. These studies show that compactness strongly depends on structured field parameterization, but they remain primarily designed for volumetric radiance fields, whose rendering is generally less direct for real-time immersive applications than explicit Gaussian rasterization.

\noindent \textbf{Compression for GS-based methods} extends compact static 3DGS representations to dynamic volumetric video, driven by advances in GS compression \cite{lu2024scaffold,hac2024,chen2025hac++,chen2025fast,sun20243dgstream,hu20254dgc,li2025gifstream} and multiview datasets \cite{joo2015panoptic,sabater2017dataset,li2022neural,gao2025bvicr,azzarelli2025vivo}. 
For static 3DGS \cite{kerbl20233dgs}, existing methods reduce redundancy through compact parameterization \cite{zhang2025mega,cho20264d}, pruning \cite{rota2024revising,hanson2025pup,lee2025optimized}, quantization \cite{wang2025compressing,navaneet2024compgs,durvasula2025contrags}, context modeling \cite{liu2024hemgs,chen2025hac++,wang2024contextgs,liu2026light4gs,zhan2025catdgs}, and rate-distortion optimization \cite{lu2024scaffold,hac2024,chen2025hac++,chen2025fast}. 
Feed-forward methods \cite{liu2025feed,chen2025fast,zhang2026d} further avoid extensive per-scene overfitting. For dynamic scenes, GS-based codecs can be broadly grouped into frame-wise predictive approaches and temporally shared representations. The former encode timestamp-dependent Gaussian sets using online updates, motion compensation, GoF-based inter-frame coding, residual prediction, or temporal pruning and quantization \cite{sun20243dgstream,gao2024hicom,hu20254dgc,zhang2026d,girish2024queen,liu2025compgspp,javed2025tc3dgs}. The latter model temporal evolution with canonical deformation, feature streams, motion-decoupled trajectories, or structured UV/attribute atlases \cite{li2025gifstream,zheng20254dgcpro,zhong20254d}. Beyond volumetric video, generative GS-based approaches have also been explored for 2D video compression \cite{liu2025exploration,teng2026gfix}.

\section{Method}
\label{sec:method}

\subsection{Preliminaries} 
\label{subsec:prelim}

\noindent \textbf{Gaussian Splatting} represents a 3D scene as learnable anisotropic Gaussian primitives $\{g_j\}_{j=1}^N$. Each primitive is parameterized as $g_j=(\bm{\mu}_j,\bm{q}_j,\bm{s}_j,\alpha_j,\bm{c}_j)$, where $\bm{\mu}_j\in\mathbb{R}^3$ is the center, $\bm{q}_j$ denotes the rotation quaternion, $\bm{s}_j$ is the scale, $\alpha_j$ is the opacity, and $\bm{c}_j$ encodes view-dependent color/appearance. Its covariance is $\bm{\Sigma}_j=\bm{R}_j\operatorname{diag}(\bm{s}_j)^2\bm{R}_j^\intercal$. Given a camera, each Gaussian is projected to a 2D elliptical footprint $\mathcal{G}_j(\bm{p})$. After depth sorting, the color values at pixel $\bm{p}$ are computed by differentiable front-to-back alpha compositing as $\bm{C}(\bm{p}) = \sum_{j=1}^{N} T_j \alpha_j \mathcal{G}_j(\bm{p}) \bm{c}_j$, where $T_j = \prod_{j'=1}^{j-1}\left(1-\alpha_{j'}\mathcal{G}_{j'}(\bm{p})\right)$ is the accumulated transmittance. This explicit differentiable formulation enables efficient rendering and high-quality novel view synthesis.

\noindent \textbf{ScaffoldGS}~\cite{lu2024scaffold} organizes 3D Gaussians with a sparse set of anchor points. Each anchor $i$ is associated with a position $\bm{x}_i$, an anchor feature $\bm{f}_i$, a learnable scaling factor $\bm{l}_i$, and $K$ learnable local offsets $\bm{O}_i=\{\bm{o}_{i,k}\}_{k=1}^{K}$. For a visible anchor, ScaffoldGS spawns $K$ neural Gaussians whose centers are given by:
\begin{equation}
\bm{\mu}_{i,k}=\bm{x}_i+\bm{o}_{i,k}\odot \bm{l}_i \label{eq:coord}
\end{equation}
The remaining Gaussian attributes, including opacity, color, rotation, and scale, are not independently stored for each spawned Gaussian. Instead, they are decoded on the fly from a view-adaptive anchor feature together with the camera-anchor distance $\delta_i$ and viewing direction $\vec{\bm{d}}_i$:
\[
\{\alpha_{i,k}\}_{k=1}^{K}=F_{\alpha}(\hat{\bm{f}}_i,\delta_i,\vec{\bm{d}}_i),\quad
\{\bm{c}_{i,k}\}_{k=1}^{K}=F_{c}(\hat{\bm{f}}_i,\delta_i,\vec{\bm{d}}_i),
\]
with $\{\bm{q}_{i,k}\}_{k=1}^{K}$ and $\{\bm{s}_{i,k}\}_{k=1}^{K}$ predicted analogously by $F_q$ and $F_s$. 
Thus, ScaffoldGS can be viewed as an anchor-conditioned neural Gaussian generator, where sparse anchors provide local context and offsets, while lightweight MLPs instantiate view-adaptive Gaussian attributes.

\subsection{Overview}
\label{subsec:overview}

\begin{figure}[t]
    \centering
    \includegraphics[width=\linewidth]{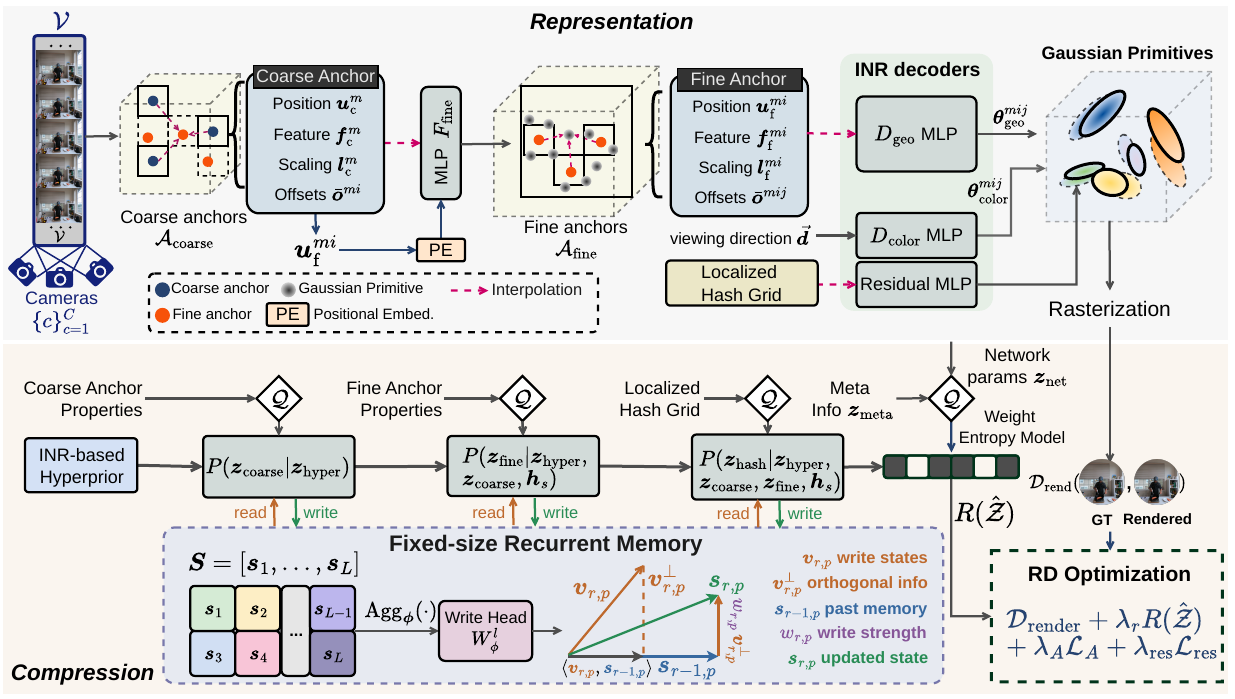}
    \caption{\textbf{Overview of SAGA.} In the representation stage, coarse anchors provide shared spatiotemporal context for fine anchors through localized kernel interpolation, while fine anchors generate Gaussian primitives via INR decoders and residual hash refinement. In the compression stage, anchors, hash grids, and neural decoder parameters are quantized and entropy coded with a hierarchical context model, where a recurrent memory bank captures long-range dependencies across sequential coding groups for improved compression efficiency.}
    \label{fig:framework}
\end{figure}
The proposed \name framework is illustrated by \autoref{fig:framework}, which consists of two major stages: representation and compression, which we describe in detail below:

\begin{figure}
    \centering
    \includegraphics[width=\linewidth]{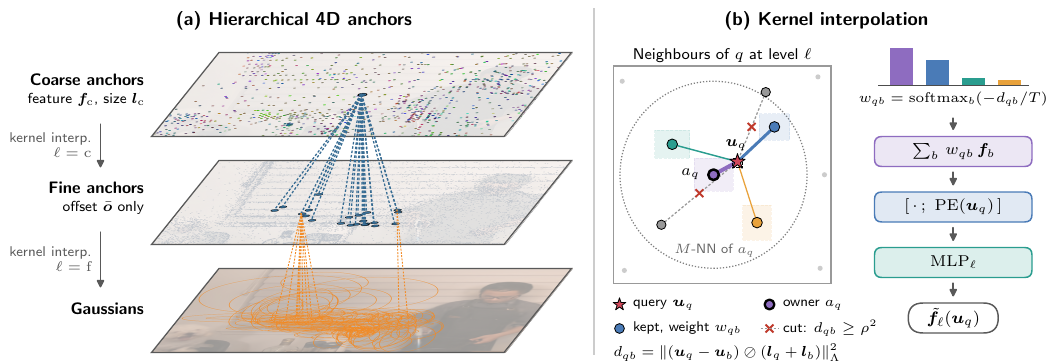}
    \caption{\textbf{Hierarchical 4D anchors and kernel interpolation.} (a) Sparse coarse anchors store features, fine anchors store only offsets, and features reach fine anchors and Gaussians by kernel interpolation; (b) a query blends its neighbours' features, weighted by support-normalized distance.}
    \label{fig:methods}
\end{figure}

\noindent\textbf{Representation.} To exploit correlations among unstructured 4D anchors, we leverage a unified \textit{interpolation-and-decoding} mechanism across the hierarchy. For each fine anchor, the neighboring coarse anchors of its parent are aggregated via kernel interpolation, as detailed in \autoref{subsec:anchor}, producing a coordinate-adaptive latent feature $\tilde{\bm{f}}_{\mathrm{c}}(\bm{u}_{\mathrm{f}}^{mi})$, which includes a positional embedding of $\bm{u}_{\mathrm{f}}^{mi}$. The fine-anchor offset $\bar{\bm{o}}^{mi}$ is part of the encoded scaffold and is used, after positional embedding, as the local coordinate input to the shared INR decoder. Specifically, $F_{\mathrm{fine}}$ predicts the remaining fine-anchor attributes as
\begin{equation}
(\bm{f}_{\mathrm{f}}^{mi},\bm{l}_{\mathrm{f}}^{mi})
=
F_{\mathrm{fine}}
\left(
\tilde{\bm{f}}_{\mathrm{c}}(\bm{u}_{\mathrm{f}}^{mi}),
\mathrm{PE}(\bar{\bm{o}}^{mi})
\right).
\end{equation}
Here $\bm{f}_{\mathrm{f}}^{mi}$ and $\bm{l}_{\mathrm{f}}^{mi}$ are the features and the support of the fine anchor, respectively. $\mathrm{PE}(\cdot)$ denotes positional embedding \cite{sitzmann2020implicit}. Gaussian primitives are generated from the fine-anchor scaffold in the same local-coordinate manner. Each fine anchor carries a set of primitive offsets $\{\bar{\bm{o}}^{mij}\}_{j=1}^{N_{mi}}$ in its local coordinate system, obtained as $\bar{\bm{o}}^{mij} = F_{\mathrm{off}}(\bm{f}_{\mathrm{f}}^{mi})_j + \bm{r}^{mij}$, where the shared predictor $F_{\mathrm{off}}$ provides a layout template and only the residuals $\bm{r}^{mij}$ are transmitted as part of the scaffold. For each primitive, we recover its 4D coordinate $\bm{u}_{\mathrm{g}}^{mij}$ from $\bar{\bm{o}}^{mij}$ and the parent fine anchor, interpolate neighboring fine anchors to obtain $\tilde{\bm{f}}_{\mathrm{f}}(\bm{u}_{\mathrm{g}}^{mij})$, again fused with $\mathrm{PE}(\bm{u}_{\mathrm{g}}^{mij})$, and decode its attributes as
\begin{equation}
\bm{\theta}_{\mathrm{geo}}^{mij}
=
F_{\mathrm{geo}}
\left(
\tilde{\bm{f}}_{\mathrm{f}}(\bm{u}_{\mathrm{g}}^{mij}),
\mathrm{PE}(\bar{\bm{o}}^{mij})
\right),
\qquad
\bm{c}_{c,s}^{mij} = F_{\mathrm{color}} \left(\tilde{\bm{f}}_{\mathrm{f}}(\bm{u}_{\mathrm{g}}^{mij}), \mathrm{PE}(\bar{\bm{o}}^{mij}), \vec{\bm{d}}_{c}^{mij}, \tau_s-\tau^{mij}
\right).
\end{equation}
Here $\bm{\theta}_{\mathrm{geo}}^{mij}$ includes opacity, spatial scaling, base rotation, temporal-opacity parameters, and polynomial motion/rotation coefficients following STG~\cite{li2024spacetime}. These coefficients are evaluated at timestamp $\tau_s$ to obtain primitive position and orientation, while $F_{\mathrm{color}}$ predicts view- and time-dependent color. Each primitive further carries a learnable binary mask $m^{mij}$~\cite{chen2025hac++}, optimized with a straight-through estimator, which gates its opacity during training and to remove/exclude invalid Gaussians. The retained primitives are rasterized to produce $\tilde{I}_{c,s}$.

\noindent \textbf{Compression.}
Parameters are quantized with learned quantization step sizes, using soft-rounding~\cite{kwan2024nvrc} during training, followed by a short quantization-noise stage. With a slight abuse of notation, we denote the tuple of quantized parameters to be entropy coded as $\hat{\mathcal{Z}}=(\bm{\hat{z}}_\mathrm{meta}, \bm{\hat{z}}_\mathrm{coarse}, \bm{\hat{z}}_{\mathrm{fine}}, \bm{\hat{z}}_{\mathrm{prim}}, \bm{\hat{z}}_{\mathrm{mask}}, \bm{\hat{z}}_{\mathrm{net}})$, including meta information, coarse/fine anchor symbols, primitive offset residuals, primitive masks, and network parameters, with the latter also covering the INR-based hyperprior~\cite{ho2026ted}. We encode $\bm{\hat{z}}_\mathrm{meta}$ independently and compress $\bm{\hat{z}}_\mathrm{net}$, which includes the decoders and the entropy model itself; the remaining symbols are entropy modeled hierarchically, as shown in \autoref{fig:framework}, conditioned on the hyperprior, previously decoded neighbors, and a memory bank. Within the coarse level, and likewise for the primitive offsets, symbols are partitioned into $K_{\mathrm c}$ parity-based coding groups according to the integer 4D cell coordinates of their anchors. For the $n$-th anchor with cell coordinate $\bm{v}_{n}^{\ell}\in\mathbb{Z}^{4}$, the group index is
\begin{equation}
    \gamma(\bm{v}_{n}^{\ell}) = \bm{\omega}^{\intercal} (\bm{v}_{n}^{\ell}\bmod 2), \qquad \bm{\omega}=(1,2,4,8)^\intercal,
\end{equation}
so that the $k$-th coding group is $\mathcal{G}_{\ell,k} = \left\{n \mid \gamma(\bm{v}_{n}^{\ell})=k \right\}$ for $k=0,\dots,K_{\mathrm c}-1$, with $K_{\mathrm c}=16$. The groups are decoded sequentially over $k$, while all symbols within the same group are decoded in parallel. Fine-anchor offsets are decoded in a single step conditioned on their parent coarse anchors. Decoding the full scaffold therefore requires $2K_{\mathrm c}+1$ sequential steps.

\subsection{Hierarchical 4D Anchor}
\label{subsec:anchor}

\noindent \textbf{Initialization and Optimization.}
We initialize the hierarchy from a unified 4D candidate pool. The sequence is split into groups of pictures (GOPs), and a short spatio-temporal Gaussian (STG)~\cite{li2024spacetime} warm-up is fitted to each GOP as a proposal stage. For every provisional Gaussian $q$, one gradient pass over the training views accumulates its screen-space gradient magnitude $S_q$ and a temporally normalized response $D_q$ that favors short-lived content. Gaussians whose $S_q$ or $D_q$ exceeds the corresponding upper quantile, and whose opacity is non-negligible, become candidates with importance
\begin{equation}
w_q =
\min\left\{
    \max\left(\frac{S_q}{Q_S}, \frac{D_q}{Q_D}\right)
    \frac{\bar{d}}{d_q},
    \; w_{\max}
\right\},
\end{equation}
where $Q_S$ and $Q_D$ are the corresponding quantile thresholds, $d_q$ is the camera distance of candidate $q$, and $\bar{d}$ is a reference camera distance used for normalization. A candidate is treated as static when it barely moves and remains visible for at least half a GOP. Static candidates are placed at the middle timestamp and de-duplicated across GOPs, while dynamic candidates are mapped to sequence time.

The resulting pool is $\mathcal{P}=\{((\bm{x}_q,\tau_q),\bm{h}_q,w_q)\}_q$, where $(\bm{x}_q,\tau_q)$, $\bm{h}_q$, and $w_q$ denote the 4D location, color, and importance, respectively. We build a greedy dyadic 4D tree using the descriptor
\begin{equation}
\bm{\phi}_q = \left[\bm{x}_q,\; \lambda_\tau \tau_q,\; \lambda_h \bm{h}_q \right],
\end{equation}
so subdivision reflects spatial, temporal, and appearance variation. The leaf with the largest importance-weighted descriptor scatter (sum of squared deviations) is split next, until the coarse-anchor budget or maximum depth is reached. A cell is split in space only, unless the weighted temporal spread within it exceeds a threshold, in which case it is also halved in time.

Leaf cells automatically become coarse anchors. For leaf $m$, $\bm{u}_{\mathrm{c}}^m$ is the cell center, $\bm{l}_{\mathrm{c}}^m$ is the cell half-size, refined during training by a learned log-scale correction, and $\bm{f}_{\mathrm{c}}^m$ is a learnable feature initialized from the weighted mean colour of its candidates. Fine-anchor capacity is allocated according to
\begin{equation}
A_m = \left(\textstyle\sum_{q\in\mathcal{P}_m} w_q\right)^\alpha (H_m+\epsilon)^\beta,
\end{equation}
where $\mathcal{P}_m$ is the candidate set in cell $m$ and $H_m$ is its importance-weighted descriptor variance. The allocation scores are converted into integer counts $N_m$ subject to  $\sum_m N_m = N_{\mathrm{budget}}, \space N_{\min} \leq N_m \leq N_{\max}$. Given $N_m$, weighted $k$-means++ seeding selects representatives, which are stored as normalized offsets, i.e., $\bar{\bm{o}}^{mi} = \left(\bm{u}^{mi}-\bm{u}_{\mathrm{c}}^m\right) \oslash \bm{l}_{\mathrm{c}}^m$,  where $\oslash$ denotes element-wise division.

\medskip
\noindent \textbf{Geometry Coding and Decoding Order.}
We use $\mathrm{c}$ and $\mathrm{f}$ to denote the coarse and fine anchor levels, respectively. Each coarse center is quantized on a $2^{B_{\mathrm{c}}}$ grid per
dimension and mapped to a 4D Morton key $\kappa_m^{\mathrm{c}}$ by interleaving the coordinate bits of $(x,y,z,\tau)$. The sorted keys are stored with Elias--Fano coding, followed by the entropy-coded cell depths and fine-anchor counts $N_m$, so the tree layout is encoded losslessly. Fine anchors are stored as offsets $\bar{\bm{o}}^{mi}$ from their parent, quantized and entropy-coded with the parent coarse anchor as context.

Attributes are decoded level by level, so each level is conditioned on the preceding one: coarse features and support deltas, then fine offsets, then Gaussian masks and offsets. Within the coarse and Gaussian levels, symbols are split into 16 groups by the parity of their 4D cell index (for a Gaussian, the local cell of its fine anchor). Group $k$ is decoded in parallel, conditioned on its decoded neighbours in groups smaller than $k$, which bounds the sequential depth to 16 steps per level.

\noindent \textbf{Kernel Interpolation.}
Let $\ell\in\{\mathrm{c},\mathrm{f}\}$ denote the source level, used to decode fine anchors from coarse anchors ($\ell=\mathrm{c}$) and Gaussian primitives from fine anchors ($\ell=\mathrm{f}$). Each source anchor $b$ has position $\bm{u}_b$, support $\bm{l}_b$, and feature $\bm{f}_b$. A query at $\bm{u}_q$ belongs to a source anchor $a_q$ (its parent or owner) and has support $\bm{l}_q$, set to $\tfrac14\bm{l}_{a_q}$ for fine anchors and $\bm{l}_{a_q}$ for Gaussians. Its candidates are the $M$ nearest 4D neighbours of $a_q$ by center distance, which the decoder recomputes for free and which also provide the entropy-coding context of the coarse and Gaussian levels. Among them, the $K$ closest with $d_{qb}<\rho^2$ form $\mathcal{N}(q)$ (always including $a_q$), and their features are blended , as illustrated in \autoref{fig:methods} (b):
\begin{equation}
\tilde{\bm{f}}_\ell(\bm{u}_q)=\operatorname{MLP}_\ell\Big(\Big[\sum_{b\in\mathcal{N}(q)}\operatorname{softmax}_b\big(-d_{qb}/T\big)\,\bm{f}_b\,;\,\mathrm{PE}(\bm{u}_q)\Big]\Big),
\quad
d_{qb}=\big\|(\bm{u}_q-\bm{u}_b)\oslash(\bm{l}_q+\bm{l}_b)\big\|_{\bm{\Lambda}}^2,
\end{equation}
where $[\cdot\,;\cdot]$ denotes concatenation, $\mathrm{PE}$ is a sinusoidal encoding, $\operatorname{MLP}_\ell$ is a two-hidden-layer ReLU MLP per level, $\oslash$ is element-wise division, and $\bm{\Lambda}=\operatorname{diag}(1,1,1,\lambda_t)$ weights the temporal term.

\subsection{Hierarchical Context Modeling}
\label{subsec:context}

Our entropy model augments the local hierarchical context with a fixed-size memory bank to capture long-range dependencies among sparse 4D anchors. Let $r$ index the sequential decoding steps induced by the hierarchy level and modulo group; for example, one step may decode the symbol blocks associated with $\mathcal{G}_{\ell,k}$, where $\ell\in\{\mathrm c,\mathrm f\}$ and $k$ is the modulo-group index. We denote the blocks decoded at step $r$ as $\mathcal{B}_r$. All blocks in $\mathcal{B}_r$ are decoded in parallel and read from the same previous memory state 
$\bm{S}_{r-1}=[\bm{s}_{r-1,1},\ldots,\bm{s}_{r-1,L}]$, where $L$ is the number of memory slots.

For each symbol block $g\in\mathcal{B}_r$, we form a memory query token $\bm{q}_g$ from its known side information and read the memory as $\bm{h}_{g}^{\mathrm{mem}}=\operatorname{Read}_{\phi}(\bm{q}_g,\bm{S}_{r-1})$. 
The entropy parameters are then predicted by 
$(\bm{\mu}_g,\bm{\sigma}_g)=E_{\phi}(\bm{e}_{g}^{\mathrm{hp}},\bm{h}_{g}^{\mathrm{parent}},\bm{h}_{g}^{\mathrm{prev}},\bm{h}_{g}^{\mathrm{mem}})$, where $\bm{e}_{g}^{\mathrm{hp}}$ is the hyperprior embedding, $\bm{h}_{g}^{\mathrm{parent}}$ is aggregated from decoded coarser-level symbols, and $\bm{h}_{g}^{\mathrm{prev}}$ uses only symbols from earlier modulo groups within the same coding stage. 
No symbol from the current batch $\mathcal{B}_r$ is used to predict another symbol in the same batch, preserving parallel entropy decoding.

After all blocks in $\mathcal{B}_r$ are decoded, their quantized values and side information are converted into decoded tokens 
$\{\tilde{\bm{z}}_g\}_{g\in\mathcal{B}_r}$ and aggregated into a batch write token 
$\bm{a}_r=\operatorname{Agg}_{\phi}(\{\tilde{\bm{z}}_g\}_{g\in\mathcal{B}_r})$. 
The memory bank contains $L$ slots, and $p\in\{1,\ldots,L\}$ indexes an individual slot. For each slot $p$, a slot-specific candidate write vector and gate are predicted from the same batch write token:
$(\bm{v}_{r,p},w_{r,p})=W_{\phi}^{p}(\bm{a}_r)$. 
To avoid repeatedly writing information already represented by the \OldMem{previous slot}, we remove from the \WriteMem{candidate write vector} the component parallel to that slot:
\begin{equation}
\WriteMem{\bm{v}_{r,p}^{\perp}}
=
\WriteMem{\bm{v}_{r,p}}
-
\frac{
\left\langle
\WriteMem{\bm{v}_{r,p}},
\OldMem{\bm{s}_{r-1,p}}
\right\rangle
}{
\|\OldMem{\bm{s}_{r-1,p}}\|_2^2+\epsilon
}
\OldMem{\bm{s}_{r-1,p}} .
\end{equation}
Each memory slot is then updated from its own previous value by combining the \OldMem{previous slot} with the \GateMem{gated} \WriteMem{orthogonal write}:
\begin{equation}
\NewMem{\bm{s}_{r,p}} = \operatorname{Norm}\left(\OldMem{\bm{s}_{r-1,p}} + \GateMem{w_{r,p}}\WriteMem{\bm{v}_{r,p}^{\perp}}\right), \qquad p=1,\ldots,L.
\end{equation}
After all $L$ slots have been updated, the new memory state is formed as $\bm{S}_{r}=[\NewMem{\bm{s}_{r,1}},\ldots,\NewMem{\bm{s}_{r,L}}]$.
Here \OldMem{$\bm{s}_{r-1,p}$} is the previous value of slot $p$, 
\WriteMem{$\bm{v}_{r,p}^{\perp}$} is the orthogonalized write content for that slot, \GateMem{$w_{r,p}$} controls the slot-specific write strength, and 
\NewMem{$\bm{s}_{r,p}$} denotes the updated slot. All slot updates are computed with respect to the previous memory state $\bm{S}_{r-1}$ and the batch-level write token $\bm{a}_r$. 

\subsection{Rate-Distortion Optimization}
\label{subsec:rd_optimization}

After the proposal warm-up and hierarchy construction, the provisional flat anchors are discarded, and only the hierarchical 4D anchor scaffold is optimized for compression. We train \name with a rate-distortion objective over the quantized transmitted symbols $\hat{\mathcal{Z}}$:
\begin{equation}
\mathcal{L} = \mathcal{D}_{\mathrm{render}} + \lambda_r R(\hat{\mathcal{Z}}) + \lambda_A\mathcal{L}_{A} + \lambda_{\mathrm{mask}}\mathcal{L}_{\mathrm{mask}}.
\end{equation}
Here $R(\hat{\mathcal{Z}})$ is the entropy-estimated bitrate under the hierarchical coding order, and $\lambda_r$, $\lambda_A$ and $\lambda_{\mathrm{mask}}$ control the relative strengths of the bitrate penalty, anchor regularization, and mask regularization terms, respectively. Quantization and the rate term are enabled after an initial reconstruction stage, with $\lambda_r$ ramped up linearly. The rendering loss follows common 3DGS practice, with $\mathcal{D}_{\mathrm{render}}=(1-\eta)\|\tilde{I}_{c,s}-I_{c,s}\|_1+\eta(1-\operatorname{SSIM}(\tilde{I}_{c,s},I_{c,s}))$, where $\tilde{I}_{c,s}$ and $I_{c,s}$ are the rendered and ground-truth frames; during training, $\tilde{I}_{c,s}$ passes through a per-camera affine color correction that is the identity at test time. The anchor regularization constrains local offsets to remain within their parent-anchor coordinate ranges. We define $\mathcal{O}_{A}=\{\bar{\bm{o}}^{mi}\}\cup\{\bar{\bm{o}}^{mij}\}$ for the coarse-to-fine and fine-to-primitive local offsets, and use $\mathcal{L}_{A}=|\mathcal{O}_{A}|^{-1}\sum_{\bar{\bm{o}}\in\mathcal{O}_{A}}\|\operatorname{ReLU}(|\bar{\bm{o}}|-\rho_A)\|_2^2$,
where $\rho_A$ is the allowed normalized boundary. This prevents the hierarchy from degenerating into unconstrained flat anchors.

\section{Experiments}

\begin{figure}[t]
    \centering
    \includegraphics[width=\linewidth]{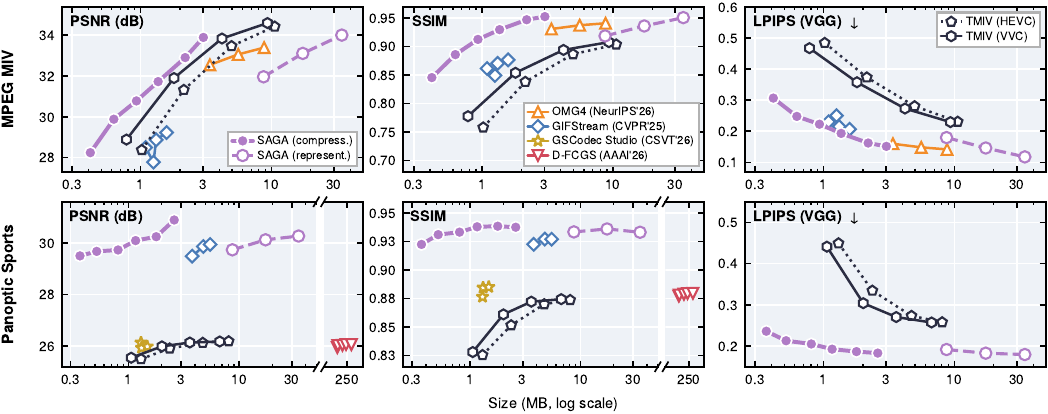}
    \caption{\textbf{RD comparison} on Panoptic Sports and MPEG MIV in terms of PSNR, SSIM and LPIPS.}
    \label{fig:rd-ssim-lpips}
\end{figure}

\begin{table*}[ht]
\centering

\begin{minipage}[t]{\textwidth}
\centering

\captionof{table}{
\textbf{Novel View Synthesis.}
We report PSNR, SSIM, LPIPS~\cite{zhang2018unreasonable}, training time in hours, decoding FPS, and model size on Neu3D and Panoptic Sports. Missing entries are denoted by ``--'' when results or implementations are unavailable. For a fair comparison, we report the \textit{raw model size} of the proposed \name, instead of the size after quantization and entropy coding.
}
\label{tab:nvs}

\resizebox{\linewidth}{!}{%
\begin{tabular}{r*{12}{c}}
\toprule
& \multicolumn{6}{c}{Neu3D~\cite{li2022neural}}
& \multicolumn{6}{c}{Panoptic Sports~\cite{Joo_2017_TPAMI}} \\
\cmidrule(lr){2-7}
\cmidrule(lr){8-13}

Method
& PSNR$\uparrow$ & SSIM$\uparrow$ & LPIPS (ALEX)$\downarrow$ & Train$\downarrow$ & FPS$\uparrow$ & Size (MB)$\downarrow$
& PSNR$\uparrow$ & SSIM$\uparrow$ & LPIPS (VGG)$\downarrow$ & Train$\downarrow$ & FPS$\uparrow$ & Size (MB)$\downarrow$ \\
\midrule

    4DGS~\cite{Wu_2024_CVPR} (CVPR'24)        & 31.57 & \trd 0.947 & 0.057 & 120 & \snd 228 & 202
                                              & 28.68 & 0.911 & \fst 0.157 & -- & -- & -- \\
    E-D3DGS~\cite{bae2024per} (ECCV'24)       & 31.42 & 0.945 & 0.037 & -- & -- & 137
                                              & 25.61 & 0.896 & \snd 0.172 & -- & 50 & 297.9 \\
    STG~\cite{li2024spacetime} (CVPR'24)      & \trd 32.05 & 0.946 & 0.050 & \snd 46 & 140 & 200
                                              & 25.09 & 0.900 & 0.181 & \fst 25 & \fst 265 & 181 \\
    MEGA~\cite{zhang2025mega} (ICCV'25)       & 31.49 & \fst 0.956 & 0.057 & -- & -- & \trd 25
                                              & -- & -- & -- & -- & -- & -- \\
    4DScaffold~\cite{cho20264d} (AAAI'26)     & 31.42 & 0.945 & \snd 0.020 & 145 & 135 & 236
                                              & 27.94 & 0.925 & 0.217 & 121 & \snd 177 & 193.7 \\
    \midrule
    SAGA (ours) [\emph{small}]                & 31.53 & 0.931 & 0.039 & \fst 34 & \fst 247 & \fst 8.71
                                              & \trd 29.74 & \trd 0.935 & 0.192 & \snd 31 & \trd 117 & \fst 8.71 \\
    SAGA (ours) [\emph{medium}]               & \snd 32.44 & 0.945 & \trd 0.021 & \trd 58 & \trd 202 & \snd 17.37
                                              & \snd 30.12 & \snd 0.936 & 0.182 & \trd 79 & 112 & \snd 17.37 \\
    SAGA (ours) [\emph{large}]                & \fst 32.81 & \snd 0.951 & \fst 0.019 & 103 & 165 & 32.54
                                              & \fst 30.43 & \fst 0.937 & \trd 0.179 & 98 & 110 & \trd 31.54 \\

\bottomrule
\end{tabular}}
\end{minipage}

\vspace{6mm}

\begin{minipage}[t]{\textwidth}
\centering

\captionof{table}{
\textbf{4D Video Compression.}
Comparison on Neu3D, Panoptic Sports, and MPEG MIV, evaluated in terms of PSNR, SSIM, LPIPS (VGG), and decoding FPS. Each BD-rate value is computed using the corresponding benchmark codec as the anchor. We only report baselines for which BD-rate can be reliably computed (i.e., with sufficient overlaps in rate-distortion range).
}
\label{tab:bdrate_comparison}

\resizebox{\linewidth}{!}{%
\begin{tabular}{r*{12}{c}}
\toprule
& \multicolumn{4}{c}{Neu3D~\cite{li2022neural}}
& \multicolumn{4}{c}{Panoptic Sports~\cite{Joo_2017_TPAMI}}
& \multicolumn{4}{c}{MPEG MIV~\cite{boyce2021mpeg}} \\
\cmidrule(lr){2-5}
\cmidrule(lr){6-9}
\cmidrule(lr){10-13}

Method
& PSNR$\uparrow$ & SSIM$\uparrow$ & LPIPS$\downarrow$ & FPS$\uparrow$
& PSNR$\uparrow$ & SSIM$\uparrow$ & LPIPS$\downarrow$ & FPS$\uparrow$
& PSNR$\uparrow$ & SSIM$\uparrow$ & LPIPS$\downarrow$ & FPS$\uparrow$ \\
\midrule

\rowcolor{lightgreen}
\multicolumn{13}{l}{\textbf{Conventional Volumetric Codec}} \\
\addlinespace[1pt]

TMIV-24.0 (HEVC, main)~\cite{boyce2021mpeg} & -- & -- & -- & 0.05 & -- & -- & -- & 0.07 & -45.89\% & -82.97\% & -85.05\% & 0.06\\

TMIV-24.0 (VVEnc, main)~\cite{boyce2021mpeg} & -- & -- & -- & 0.03 & -- & -- & -- & 0.05 & -18.41\% & -75.24\% & -72.27\% & 0.05 \\

\addlinespace[2pt]
\rowcolor{lightred}
\multicolumn{13}{l}{\textbf{GS-based Methods}} \\
\addlinespace[2pt]

4DGC~\cite{hu20254dgc} (CVPR'25) & -62.39\% & -41.59\% & -61.35\% & 110 & -- & -- & -- & -- & -32.38\% & -36.13\% & -42.61\% & 115 \\

GIFStream~\cite{li2025gifstream} (CVPR'25) & -90.88\% & -81.72\% & -69.55\% & 91 & -84.87\% & -87.06\% & -76.95\% & 99 & -67.09\% & -67.10\% & -45.87\% & 67 \\

TED-4DGS~\cite{ho2026ted} (WACV'26) & -64.30\% & -- & -- & -- & -- & -- & -- & -- & -- & -- & -- & -- \\

IC-4DGS~\cite{yang2026ic} (ICIP'26) & -22.62\% & -- & -- & -- & -- & -- & -- & -- & -- & -- & -- & -- \\

D-FCGS~\cite{zhang2026d} (AAAI'26) & -99.31\% & -99.08\% & -92.05\% & 4.9 & -- & -- & -- & 6.7 & -- & -- & -- & -- \\


OMG4~\cite{Lee_Optimized_NeurIPS_2026} (NeurIPS'26) & -65.14\% & -17.01\% & -24.28\% & 245 & -- & -- & -- & -- & -- & -- & -- & -- \\

\addlinespace[3pt]
\hdashline
\addlinespace[3pt]

\textbf{\name{} (ours)}
& \textbf{0.00\%} & \textbf{0.00\%} & \textbf{0.00\%} & 85 & \textbf{0.00\%} & \textbf{0.00\%} & \textbf{0.00\%} & -- & \textbf{0.00\%} & \textbf{0.00\%} & \textbf{0.00\%} & 97 \\
\bottomrule
\end{tabular}%
}

\end{minipage}

\end{table*}

\subsection{Experimental Settings}

\noindent \textbf{Datasets.} We conduct experiments on three datasets: the widely used Neu3D \cite{li2022neural}, Panoptic Sports \cite{Joo_2017_TPAMI}, and the multi-view texture-and-depth video sequences defined in the MPEG Immersive Video (MIV) Common Test Conditions \cite{boyce2021mpeg}. Neu3D consists of 6 indoor dynamic multi-view sequences captured by 18-21 cameras at a resolution of 2704$\times$2028. Following prior works \cite{li2025gifstream,zhang2026d}, we represent and compress sequences in Neu3D at \textit{half resolution} for the first 300 frames \cite{li2025gifstream}. Panoptic Sports \cite{Joo_2017_TPAMI} contains six dynamic sports sequences at 640$\times$360 resolution, each recorded by 31 cameras; following \cite{li2025gifstream}, we report results on the \textit{basketball} and \textit{boxes} sequences. For each sequence, cameras 0, 10, 15, and 30 are used for testing, while the remaining cameras are utilized for training according to the protocol in \cite{luiten2024dynamic}. The MPEG TMIV CTC dataset comprises 18 scenes captured by 9-16 cameras at 1080p resolution, from which we select the first 65 frames of mandatory sequences, as specified by the MPEG MIV CTC (see details in \textbf{Appendix}\ref{app:dataset}), for benchmarking. 

\noindent \textbf{Baselines.} For novel view synthesis, we compare \name against state-of-the-art compact GS-based methods, including 4DGS~\cite{Wu_2024_CVPR}, STG~\cite{li2024spacetime}, E-D3DGS~\cite{bae2024per}, MEGA~\cite{zhang2025mega}, 4DScaffold~\cite{cho20264d}. For volumetric video compression, we evaluate \name against a broad range of approaches, including the \ul{conventional} MPEG immersive video test model TMIV-24.0~\cite{boyce2021mpeg} in the \textit{MIV Main} configuration, using HM-18.0 and VVEnc-1.12.0 for texture and geometry atlas coding. We use the provided source-view depths for MIV-CTC and estimate depths for Neu3D and Panoptic Sports using Depth-Anything-3~\cite{lin2025depth3}, aligning the depth scales with the camera calibration (see details in \textbf{Appendix}~\ref{app:baseline_config}). We also compare against \ul{GS-based compression} methods, including 4DGC~\cite{hu20254dgc}, GIFStream~\cite{li2025gifstream}, TED-4DGS~\cite{ho2026ted}, IC-4DGS~\cite{yang2026ic}, D-FCGS~\cite{zhang2026d}, GSCodec Studio~\cite{li2026gscodec}, and OMG4~\cite{Lee_Optimized_NeurIPS_2026}. It is noted that we report BD-rate results only for methods whose rate-distortion ranges overlap sufficiently with ours to support reliable computation, which, as a consequence, excludes GSCodec Studio and the FTGS variant of OMG4 from \autoref{tab:bdrate_comparison}.

\noindent \textbf{Metrics.} We use PSNR, SSIM, and LPIPS \cite{zhang2018unreasonable} to measure the reconstruction quality of the synthesized multiviews. For 4D volumetric compression, we report the Bj{\o}ntegaard Delta rate (BD-rate) \cite{bdrate} to quantify the relative compression efficiency between different codecs.

\subsection{Experimental Results}

\begin{figure}[t]
    \centering
    \includegraphics[width=\linewidth]{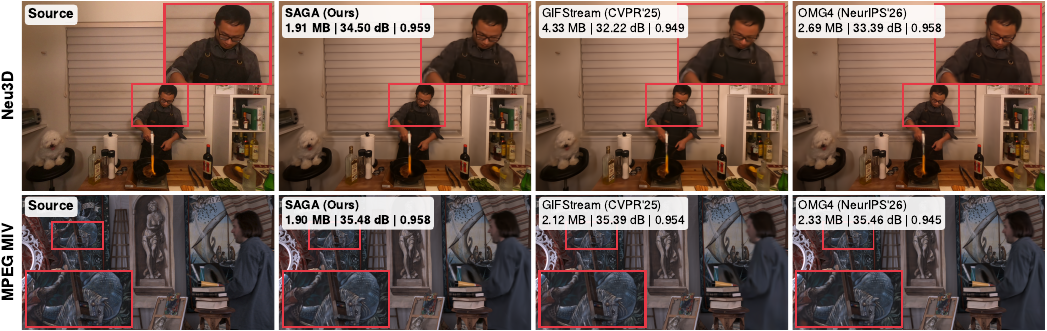}
    \caption{\textbf{Exemplary qualitative comparisons} on Neu3D (\textbf{top}) and MPEG MIV (\textbf{bottom}).}
    \label{fig:qualitative}
\end{figure}

\noindent \textbf{Novel View Synthesis.}
As summarized in \autoref{tab:nvs}, we first evaluate \name as a compact dynamic GS representation before quantization and entropy coding. Across Neu3D and Panoptic Sports, \name provides a favorable fidelity-size trade-off, with its small variant achieving the smallest reported raw model size among compared methods. On Neu3D, the medium variant already surpasses all evaluated baselines in PSNR with a 17.37\,MB model, while the large variant further achieves the best PSNR and LPIPS at 32.81\,dB and 0.019, respectively. On Panoptic Sports, even the small variant outperforms all evaluated baselines in PSNR and SSIM, using only 8.7\,MB. Increasing model capacity further improves reconstruction quality, although the LPIPS remains higher than that of 4DGS and E-D3DGS. These results show that the proposed spacetime anchor hierarchy and inter-anchor decoding provide a compact representation with competitive reconstruction quality before entropy coding; additional gains from quantization and entropy modeling are evaluated in \autoref{tab:bdrate_comparison}.

\noindent \textbf{Volumetric Video Compression.}
As summarized in \autoref{tab:bdrate_comparison}, \name achieves consistent rate-distortion improvements over the evaluated GS-based volumetric codecs across the available comparisons. Compared with GIFStream~\cite{li2025gifstream}, \name obtains large BD-rate reductions across all three datasets, with at least 67.09\%, 67.10\%, and 45.87\% savings measured by PSNR, SSIM, and LPIPS, respectively. The gains also extend to recent baselines on Neu3D, with PSNR BD-rate savings exceeding 60\% over both TED-4DGS and OMG4, indicating that the compression advantage is not limited to a single baseline. Beyond GS-based methods, \name also outperforms both TMIV configurations on MPEG MIV. Even against the VVEnc configuration, it achieves a PSNR BD-rate reduction of 18.41\%, with larger savings of over 70\% under SSIM and LPIPS. Together, these results demonstrate improved compression efficiency under both distortion-based and perceptual quality metrics. The corresponding rate-distortion curves and qualitative comparisons are provided in \autoref{fig:rd-ssim-lpips} and \autoref{fig:qualitative}, with the latter further illustrating that \name preserves finer texture details at lower bitrates than GIFStream and OMG4.

\noindent \textbf{Complexity and Rate Distribution.}
We measure training cost in GPU-hours and decoding speed in FPS on a single NVIDIA RTX 3090 GPU, and report the component-wise bitrate and runtime breakdown of \name in \autoref{tab:ablation} \textbf{(a)}. As summarized in \autoref{tab:nvs} and \autoref{tab:bdrate_comparison}, the compression gains of \name are achieved while retaining real-time decoding, reaching 85 FPS on Neu3D and 97 FPS on MPEG MIV. Its decoding throughput is comparable to GIFStream on Neu3D and higher on MPEG MIV, although faster methods such as OMG4 offer a different trade-off between compression efficiency and decoding speed. Training remains a notable cost: the small variant requires less training time than 4DScaffold on both datasets, but more than STG. In terms of rate distribution (Table~\ref{tab:ablation}~(a)), quantization and entropy coding reduce the anchor scaffold (coarse anchors, fine anchors and Gaussian offsets) from 33.87\,MB to 1.56\,MB, a 95.4\% reduction, with the largest gain on the Gaussian offsets ($43\times$). Fine anchors and Gaussian offsets nevertheless still account for about 69\% of the coded bitstream, indicating that scaffold coding remains the main contributor to the total bitrate.

\subsection{Ablation Study}

\begin{table*}[t]
  \centering
  \caption{\textbf{Ablation, memory-slot analysis, and bitstream breakdown on Neu3D.} BD-rate is computed in terms of PSNR using SAGA as the anchor.}
  \label{tab:ablation}

  \begin{minipage}[t]{0.33\linewidth}
    \centering
    \scriptsize
    \textbf{(a) Bitstream}\\[2pt]
    \setlength{\tabcolsep}{2.5pt}
    \begin{tabular}{rcc}
      \toprule
      Component & \makecell{Before EC\\MB (\%)} & \makecell{After EC\\MB (\%)} \\
      \midrule
      Coarse anc.    & 3.66 (10.6)  & 0.259 (13.7) \\
      Fine anc.      & 2.75 (8.0)   & 0.672 (35.4) \\
      Gauss. offsets & 27.47 (79.5) & 0.633 (33.4) \\
      Gauss. mask    & --           & 0.116 (6.1)  \\
      MLPs           & 0.67 (1.9)   & 0.086 (4.5)  \\
      Entropy        & --           & 0.021 (1.1)  \\
      Side info      & --           & 0.111 (5.8)  \\
      \midrule
      Total          & 34.54        & 1.897        \\
      \bottomrule
    \end{tabular}
  \end{minipage}%
  \hfill
  \begin{minipage}[t]{0.65\linewidth}
    \centering

    \begin{minipage}[t]{0.56923\linewidth}
      \centering
      \scriptsize
      \textbf{(b) Component ablation}\\[2pt]
      \setlength{\tabcolsep}{2.5pt}
      \resizebox{\linewidth}{!}{%
        \begin{tabular}{lcccccc}
          \toprule
          Ver. & C/F & Interp. & Res. & MLP-C & Upd. & BD-rate$\downarrow$ \\
          \midrule
          V1   & \xmark & \cmark & \cmark & \cmark & Orth.     & 13.04\% \\
          V2   & \cmark & \xmark & \cmark & \cmark & Orth.     & 11.24\% \\
          V3   & \cmark & \cmark & \xmark & \cmark & Orth.     & 1.03\%  \\
          V4   & \cmark & \cmark & \cmark & \xmark & Orth.     & 7.22\%  \\
          V5   & \cmark & \cmark & \cmark & \cmark & Non-orth. & 8.17\%  \\
          \midrule
          SAGA & \cmark & \cmark & \cmark & \cmark & Orth.     & 0.0\%   \\
          \bottomrule
        \end{tabular}%
      }
    \end{minipage}%
    \hfill
    \begin{minipage}[t]{0.4\linewidth}
      \centering
      \scriptsize
      \textbf{(c) Memory slots}\\[2pt]
      \setlength{\tabcolsep}{2.5pt}
      \begin{tabular}{lcc}
        \toprule
        \# Slots & BD-rate$\downarrow$ & FPS$\uparrow$ \\
        \midrule
        32           & +8.87\%  & 93.1 \\
        64 (default) & 0.0\%    & 85.3 \\
        128          & -10.27\% & 68.9 \\
        256          & -15.66\% & 50.3 \\
        512          & -16.19\% & 36.6 \\
        \bottomrule
      \end{tabular}
    \end{minipage}%

    \par\vspace{4pt}

    \begin{minipage}{\linewidth}
      \raggedright
      \scriptsize
      C/F: coarse-to-fine scaffold;
      Interp.: inter-anchor interpolation;
      Res.: residual hash;
      MLP-C: MLP compression;
      Upd.: memory update.
      Orth.: orthogonality-guided;
      Non-orth.: gatedresidual without orthogonal projection.
    \end{minipage}

  \end{minipage}
\end{table*}

To investigate different design components in \name, we created the following variants: V1 removes the coarse-to-fine scaffold and uses flat anchors; V2 disables inter-anchor feature interpolation and decodes finer representations only from its own anchor; V3 removes the residual hash grid; V4 disables MLP compression; and V5 replaces the proposed orthogonality-guided memory update with a non-orthogonal gated residual update. The full model is used as the anchor for BD-rate computation. As shown in \autoref{tab:ablation} \textbf{(b)}, each ablated variant incurs BD-rate loss, demonstrating the effectiveness of the corresponding component. Removing the hierarchy or inter-anchor interpolation substantially degrades RD performance, supporting the importance of both components, while the 8.17\% loss of V5 further confirms the contribution of the orthogonalized memory update.
 
\noindent \textbf{Impact of Memory Slot Size.}
We also ablate the impact of memory slot size on compression performance and decoding speed, as shown in \autoref{tab:ablation} \textbf{(c)}. It can be observed that naively increasing the memory slot size offers diminishing returns in compression efficiency; this indicates that the long-context spatiotemporal dependency is already well captured with fewer memory slots, i.e., no more than 256. We hence choose 64 slots as the default for real-time decoding, while larger memories provide better compression at a substantial throughput cost. 

\section{Conclusion}

In this paper, we presented \textbf{\name}, a GS-based volumetric video compression framework built on hierarchical sparse 4D anchor scaffolds, which exploit inter-anchor redundancy and model local dynamics at variable granularity. Sparse coarse anchors decode fine anchors, which instantiate Gaussian primitives via learned offsets, and learnable masks prune redundant primitives. A long-context entropy model further summarizes decoded contexts into bounded memory for efficient entropy-parameter prediction. Experiments on Neu3D, Panoptic Sports, and MPEG MIV CTC benchmarks demonstrate that \name outperforms existing GS-based codecs in compression efficiency, with average BD-rate reductions (measured by PSNR) of 64.30\%, 65.14\%, and 90.88\%, over TED-4DGS, OMG4, and GIFStream on the Neu3D testset, respectively. 

\noindent \textbf{Limitations and Future Work.} \name{} still requires per-scene optimization, which may incur notable encoding time for long or complex sequences, and fast motion, topology changes, and sparse camera coverage may also challenge reconstruction. Future work will explore faster encoding, adaptive anchor allocation, and standardized bitstreams for scalability, random access, and deployment.

\begin{ack}
This work was funded by the UKRI MyWorld Strength in Places Programme (SIPF00006/1) and the Royal Society International Exchanges programme (IEC\textbackslash{}R3\textbackslash{}253264). We thank Mr He-Bi Yang for providing the RD numbers for TED-4DGS and IC-4DGS. We thank the University of Bristol Advanced Computing Research Centre and Isambard-AI for providing computational resources. Isambard-AI is funded by the UK Government via UKRI and STFC [ST/AIRR/I-A-I/1023].
\end{ack}

{\small
\bibliographystyle{abbrv}
\bibliography{main}
}

\clearpage
\appendix

\setcounter{table}{0}
\renewcommand{\thetable}{\thesection.\arabic{table}}

\setcounter{figure}{0}
\renewcommand{\thefigure}{\thesection.\arabic{figure}}

\section{Broader Impact}
This work improves the efficiency of volumetric video compression, with potential benefits for telepresence, remote collaboration, education, cultural preservation, digital media production, and immersive communication. By reducing storage and transmission costs while retaining real-time decoding, \name may make high-quality immersive media more accessible under limited bandwidth and compute budgets, and may reduce the infrastructure and energy costs of large-scale volumetric video delivery.

At the same time, more compact and realistic volumetric representations may increase risks related to privacy, unauthorized capture, surveillance, and deceptive or non-consensual synthetic media, especially when the content contains identifiable people or private environments. Responsible deployment should therefore consider consent, access control, provenance tracking, watermarking or authentication, and dataset governance. Our work focuses on compression and reconstruction quality on established research datasets, and does not develop tools for identity manipulation, text-driven content synthesis, or surveillance.

\section{Dataset/Code License}

We list the licenses of all datasets and the code used in this paper in \autoref{tab:assets}.

\begin{table}[h]
  \centering
  \caption{\textbf{Existing assets used in this work.} We list the datasets, codebases, codecs, and models used in our experiments, together with their sources and license or usage terms where available.}
  \label{tab:assets}
  \resizebox{\linewidth}{!}{%
    \begin{tabular}{llll}
      \toprule
      Asset & Type & Source / Citation & License / Terms \\
      \midrule
      Neu3D
        & Dataset
        & \cite{li2022neural}
        & CC BY-NC 4.0 \\
      Panoptic Sports / Panoptic Studio
        & Dataset
        & \cite{Joo_2017_TPAMI}
        & Research-only, non-commercial \\
      MPEG MIV CTC sequences
        & Dataset
        & \cite{boyce2021mpeg}
        & Per-sequence licenses (MIV content database) \\
      TMIV-24.0
        & Codec / software
        & \cite{boyce2021mpeg}
        & BSD 3-Clause \\
      HM-18.0
        & Codec / software
        & \cite{sullivan2012overview}
        & BSD 3-Clause \\
      VVenC
        & Codec / software
        & \cite{bross2021overview}
        & BSD 3-Clause Clear (no patent grant) \\
      Depth Anything 3 (DA3-Large)
        & Code / model weights
        & \cite{lin2025depth3}
        & DA3-Large weights: CC BY-NC 4.0 \\
      3DGStream
        & Code / method
        & \cite{sun20243dgstream}
        & MIT \\
      4D Scaffold GS
        & Code / method
        & \cite{cho20264d}
        & Gaussian-Splatting License (non-commercial) \\
      4DGC
        & Code / method
        & \cite{hu20254dgc}
        & No explicit license \\
      STG
        & Code / method
        & \cite{li2024spacetime}
        & MIT; bundled 3DGS under Gaussian-Splatting License \\
      GIFStream
        & Code / method
        & \cite{li2025gifstream}
        & Scaffold-GS parts under Gaussian-Splatting License \\
      D-FCGS
        & Code / method
        & \cite{zhang2026d}
        & Not specified \\
      OMG4
        & Code / model weights
        & \cite{Lee_Optimized_NeurIPS_2026}
        & No explicit license \\
      GSCodec Studio
        & Code / method
        & \cite{li2026gscodec}
        & Apache-2.0 \\
      TED-4DGS
        & Code / method
        & \cite{ho2026ted}
        & No explicit license \\
      IC-4DGS
        & RD results
        & \cite{yang2026ic}
        & RD results provided by the authors \\
      LPIPS
        & Evaluation metric / code
        & \cite{zhang2018unreasonable}
        & BSD 2-Clause \\
      \bottomrule
    \end{tabular}%
  }
\end{table}

\section{Detailed Configurations}

\subsection{SAGA Configurations} 
In \autoref{tab:config} we provide the detailed model configurations for the proposed \name.

\begin{table}[ht]
\centering
\caption{\textbf{Model and training configurations.} Model-size-dependent entries are listed in S / M / L order; all other settings are shared.}
\label{tab:config}
\small
\setlength{\tabcolsep}{5pt}
\renewcommand{\arraystretch}{1.08}
\begin{tabularx}{\linewidth}{@{}lXc@{}}
\toprule
\textbf{Module} & \textbf{Configuration} & \textbf{Value (S / M /L)} \\
\midrule

\module{Hierarchical 4D anchor scaffold}
& \subcfg{Coarse-anchor budget} & 6k / 12k / 24k \\
& \subcfg{Fine-anchor budget, $N_{\mathrm{budget}}$} & 45k / 90k / 180k \\
& \subcfg{Fine anchors per coarse anchor, $N_{\min}/N_{\max}$} & 1 / 32 \\
& \subcfg{Maximum tree depth} & 14 \\
& \subcfg{Temporal and statistic scaling, $\lambda_{\tau}, \lambda_{\bm h}$} & 6.0 / 0.1 \\
& \subcfg{Anchor-allocation exponents, $\alpha,\beta$} & 0.5 / 0.5 \\
& \subcfg{Local-offset boundary, $\lambda_A$} & 1.0 / 0.01 \\

\arrayrulecolor{gray!45}\hdashline\arrayrulecolor{black}

\module{Kernel interpolation}
& \subcfg{4D kNN candidates and neighbors, $M,K$} & 12 / 8 \\
& \subcfg{Support radii, $\rho_{\mathrm c}, \rho_{\mathrm f}$} & 3.0 / 3.0 \\
& \subcfg{Kernel temperatures, $T_{\mathrm c}, T_{\mathrm f}$} & 1.0 / 1.0 \\
& \subcfg{Temporal distance weight, $\lambda_t$} & 1.0 \\
& \subcfg{Interpolation MLP width, $\operatorname{MLP}_{\mathrm c},\operatorname{MLP}_{\mathrm f}$ (depth 3)} & 64 / 96 / 128 \\

\arrayrulecolor{gray!45}\hdashline\arrayrulecolor{black}

\module{Primitive mask}
& \subcfg{Mask threshold on $\operatorname{sigmoid}$ logit} & 0.01 \\
& \subcfg{Mask loss weight, $\lambda_{\mathrm{mask}}$} & $1.0\times10^{-3}$ \\

\arrayrulecolor{gray!45}\hdashline\arrayrulecolor{black}

\module{Hierarchical entropy model}
& \subcfg{Coarse / fine-anchor / primitive coding groups} & 16 / 1 / 16 \\
& \subcfg{INR hyperprior depth / width} & 3 / 16 \\
& \subcfg{Memory slots and slot dimension, $L$} & 64 / 16 \\
& \subcfg{Memory read (single-head attention) dimension} & 16 \\
& \subcfg{Entropy-parameter network depth / width} & 3 / 16 \\
& \subcfg{Batch-token embedding depth / width, aggregation} & 3 / 16, mean \\

\arrayrulecolor{gray!45}\hdashline\arrayrulecolor{black}

\module{Optimization and coding}
& \subcfg{Proposal warm-up (per 50-frame GOP) / compression iterations} & 4k / 30k \\
& \subcfg{Quantization start / rate warm-up / quant-noise stage} & 10k / 2k / last 3k \\
& \subcfg{Initial quantization precision} & 8 bits \\
& \subcfg{Rate weight $\lambda_r$ (per MB)} & 0.0003, 0.004, 0.015, 0.06, 0.001 \\
& \subcfg{Optimizer and initial learning rate} & Adam / $10^{-3}$--$10^{-2}$ \\
& \subcfg{Learning-rate schedule} & Exp.\ decay to $1\%$ \\
& \subcfg{Rendering SSIM weight, $\eta$} & 0.2 \\

\bottomrule
\end{tabularx}
\end{table}

\subsection{Dataset Preparation} \label{app:dataset}

\noindent \textbf{MPEG MIV.} Following the MPEG Immersive Video Common Test Conditions (CTC)~\cite{boyce2021mpeg}, we select five sequences from the MIV main-anchor configuration: \textit{ClassroomVideo (A01)}, \textit{Chess (B02)}, \textit{Painter (D01)}, \textit{Frog (E01)}, and \textit{Carpark (E02)}. These sequences comprise two computer-generated and three natural scenes, covering diverse scene content and camera configurations, including omnidirectional (equirectangular) and perspective source views with 9 to 16 cameras. We use the first 65 frames of each sequence. This selection follows the standardized CTC benchmark to evaluate our method on representative multi-view video content.

\subsection{Baseline Configurations} \label{app:baseline_config}

\noindent \textbf{MIV Test Model.} The MPEG Immersive Video Test Model (TMIV)~\cite{boyce2021mpeg} is the reference software for MPEG Immersive Video. It compresses multi-view immersive content by exploiting inter-view redundancy and packing pruned texture and geometry information into video atlases. We use TMIV 24.0 with the MIV Main anchor configuration, in which depth maps are required as geometry inputs and are compressed together with texture; we therefore report the size of the complete multiplexed MIV
bitstream, which contains the texture and geometry atlases as well as the atlas metadata. For the MPEG MIV sequences, we use the first 65 frames of A01, B02, D01, E01 and E02 with the supplied ground-truth depth maps and the anchor QPs of the common test conditions. For Neu3D and Panoptic Sports, RGB frames are converted to 10-bit YUV 4:2:0 (BT.709, limited range) and texture and geometry are coded with QPs $\{28,34,40,46\}$ and $\{6,10,16,22\}$, respectively, for RP1--RP4. We invoke TMIV through its \texttt{encode.py} script, which runs the TMIV encoder, the selected video encoder for each atlas component, the TMIV multiplexer, and the parser/bitrate report.

\begin{lstlisting}[
  language=bash,
  breaklines=true,
  columns=fullflexible,
  keepspaces=true
]
python3 <tmiv_root>/scripts/encode.py \
  -i <input_yuv_root> \
  -o <output_root> \
  -s <content_id> \
  -n <frame_count> \
  -f 0 \
  -c <tmiv_root>/config/ctc/miv_main_anchor/A_1_TMIV_encode.json \
  -m <tmiv_root>/config/ctc/miv_main_anchor/A_3_TMIV_mux.json \
  -q <qp_file> \
  -v VVenC \
  -C <tmiv_install_dir>/share/config/vvenc/randomaccess_medium.cfg \
  -r <rate_point> \
  -t <tmiv_install_dir> \
  --config-dir <tmiv_config_dir> \
  -j <threads>
\end{lstlisting}

The synthesized test views used for distortion evaluation are decoded from the TMIV bitstream, while the best-reference views are rendered directly from the source views for metric computation:

\begin{lstlisting}[
  language=bash,
  breaklines=true,
  columns=fullflexible,
  keepspaces=true
]
<tmiv_install_dir>/bin/TmivDecoder \
  -c <tmiv_root>/config/ctc/miv_main_anchor/A_4_TMIV_decode.json \
  -p configDirectory <tmiv_config_dir> \
  -p inputDirectory <output_root> \
  -p outputDirectory <output_root> \
  -s <content_id> -n <frame_count> -N <output_frame_count> \
  -r <rate_point> -P <pose_trace_id> -j <threads>

<tmiv_install_dir>/bin/TmivRenderer \
  -c <tmiv_root>/config/ctc/best_reference/R_1_TMIV_render.json \
  -p configDirectory <tmiv_config_dir> \
  -p inputDirectory <input_yuv_root> \
  -p outputDirectory <output_root> \
  -s <content_id> -n <frame_count> -N <output_frame_count> \
  -f 0 -r RP0 -P <pose_trace_id> -j <threads>
\end{lstlisting}

Internally, when VVenC is selected, each atlas video component is encoded by a command of the following form:

\begin{lstlisting}[
  language=bash,
  breaklines=true,
  columns=fullflexible,
  keepspaces=true
]
vvencFFapp \
  -c <video_encoder_config_file> \
  -i <atlas_yuv_file> \
  -b <video_bitstream_file> \
  -s <width>x<height> \
  -q <qp> \
  -f <frame_count> \
  -fr <frame_rate> \
  --InputBitDepth 10 \
  --OutputBitDepth 10 \
  --CbQpOffset <chroma_qp_offset> \
  --CrQpOffset <chroma_qp_offset> \
  --Threads <threads>
\end{lstlisting}

\noindent \textbf{Depth Estimation.} Neu3D and Panoptic Sports do not provide ground-truth depth. We estimate per-view depth maps using Depth Anything 3 (DA3-Large)~\cite{lin2025depth3}, running pose-conditioned multi-view inference over all source views of each timestamp so that the predicted depths share the scale of the given camera extrinsics:
\begin{lstlisting}[
  language=Python,
  breaklines=true,
  columns=fullflexible,
  keepspaces=true
]
pred = model.inference(
    image=[view["images"][t] for view in source_views],
    intrinsics=intrinsics,
    extrinsics=extrinsics,
    align_to_input_ext_scale=True,
    process_res=504,
    process_res_method="upper_bound_resize",
)
\end{lstlisting}
The held-out test views (Neu3D \texttt{cam00} and the Panoptic test cameras)
are excluded from inference. The predicted depths are resampled to the texture
resolution with nearest-neighbour sampling and quantized to 16-bit inverse
depth, with a fixed near/far range per sequence (0.9$\times$ the minimum and
1.1$\times$ the maximum predicted depth); the value 0 is reserved for invalid
samples:
\begin{lstlisting}[
  language=Python,
  breaklines=true,
  columns=fullflexible,
  keepspaces=true
]
inv = (
    (1 / np.clip(z, near, far) - 1 / far)
    / (1 / near - 1 / far)
)
q = np.clip(np.rint(inv * 65535), 1, 65535).astype(np.uint16)
\end{lstlisting}

\section{Additional Experimental Results}

\subsection{Bitstream and Inference Speed Breakdown}

\begin{figure}
    \centering
    \includegraphics[width=0.7\linewidth]{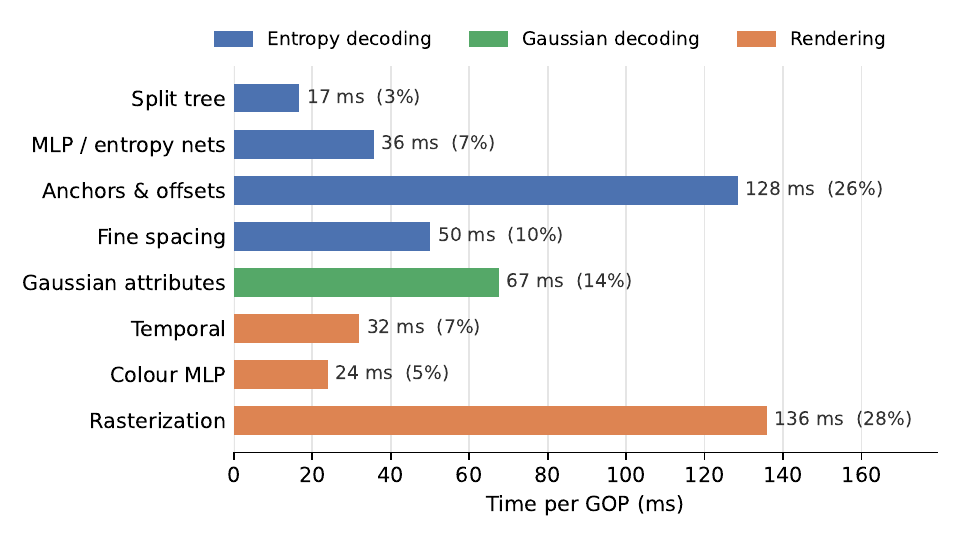}
    \caption{\textbf{Inference speed breakdown}.}
    \label{fig:app-infernece-breakdown}
\end{figure}

\begin{figure}
    \centering
    \includegraphics[width=0.8\linewidth]{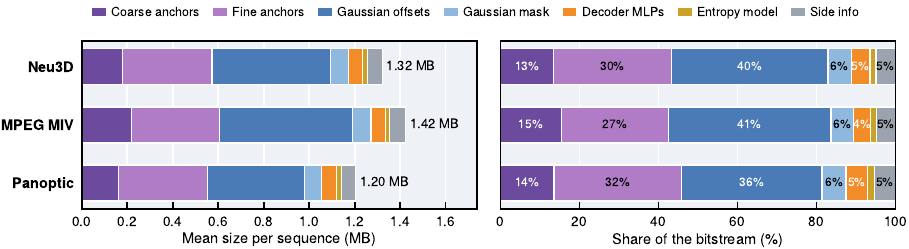}
    \caption{\textbf{Bitstream  breakdown}.}
    \label{fig:app-bit-breakdown}
\end{figure}

\noindent\textbf{Inference speed breakdown.}
As shown in Fig.~\ref{fig:app-infernece-breakdown}, rasterization (28\%) and anchor/offset decoding (26\%) dominate the inference time, accounting for more than half of the total runtime. In contrast, entropy decoding and other lightweight operations introduce relatively limited overhead.

\noindent\textbf{Bitstream breakdown.} As shown in Fig.~\ref{fig:app-bit-breakdown}, Gaussian-related parameters account for the majority of the bitstream across different datasets, with Gaussian offsets contributing the
largest share (approximately 41--46\%). This suggests that Gaussian parameter representation, particularly offsets, remains the primary bottleneck for further compression.

\subsection{Raw Values}
In \autoref{tab:rd_results_appendix} we provide the per-sequence raw values for the RD-curves of \name on the Neu3D and MPEG MIV benchmark.

\begin{table*}[ht]
\centering
\caption{\textbf{Rate-distortion results on Neu3D and MPEG MIV.}}
\label{tab:rd_results_appendix}
\scriptsize
\setlength{\tabcolsep}{3.5pt}
\renewcommand{\arraystretch}{1.08}

\begin{minipage}{\linewidth}
\centering
\textbf{(a) Neu3D}
\vspace{0.35em}

\begin{tabular}{rccccc@{\hspace{1.0em}}rccccc}
\toprule
\textbf{Sequence} & \textbf{QP} & \textbf{Size (MB)} & \textbf{PSNR} & \textbf{SSIM} & \textbf{LPIPS}
&
\textbf{Sequence} & \textbf{QP} & \textbf{Size (MB)} & \textbf{PSNR} & \textbf{SSIM} & \textbf{LPIPS} \\
\midrule

\multirow{5}{*}{\texttt{coffee\_martini}}
& $\mathrm{QP}_1$ & 2.625 & 27.476 & 0.9092 & 0.1331
&
\multirow{5}{*}{\texttt{flame\_salmon}}
& $\mathrm{QP}_1$ & 2.516 & 27.655 & 0.9074 & 0.1135 \\
& $\mathrm{QP}_2$ & 2.063 & 27.331 & 0.8931 & 0.1391
&
& $\mathrm{QP}_2$ & 2.933 & 28.207 & 0.9165 & 0.0917 \\
& $\mathrm{QP}_3$ & 1.775 & 27.451 & 0.8983 & 0.1240
&
& $\mathrm{QP}_3$ & 1.843 & 28.084 & 0.9114 & 0.1030 \\
& $\mathrm{QP}_4$ & 1.379 & 27.026 & 0.8851 & 0.1564
&
& $\mathrm{QP}_4$ & 1.433 & 28.056 & 0.9094 & 0.1048 \\
& $\mathrm{QP}_5$ & 1.020 & 26.390 & 0.8717 & 0.1859
&
& $\mathrm{QP}_5$ & 1.092 & 27.137 & 0.8875 & 0.1457 \\
\midrule

\multirow{5}{*}{\texttt{cook\_spinach}}
& $\mathrm{QP}_1$ & 4.259 & 33.956 & 0.9565 & 0.0450
&
\multirow{5}{*}{\texttt{flame\_steak}}
& $\mathrm{QP}_1$ & 2.613 & 33.864 & 0.9574 & 0.0477 \\
& $\mathrm{QP}_2$ & 3.255 & 33.742 & 0.9558 & 0.0460
&
& $\mathrm{QP}_2$ & 3.035 & 34.462 & 0.9603 & 0.0386 \\
& $\mathrm{QP}_3$ & 1.837 & 33.110 & 0.9504 & 0.0583
&
& $\mathrm{QP}_3$ & 1.909 & 34.496 & 0.9587 & 0.0430 \\
& $\mathrm{QP}_4$ & 1.336 & 32.703 & 0.9467 & 0.0642
&
& $\mathrm{QP}_4$ & 1.241 & 33.748 & 0.9532 & 0.0564 \\
& $\mathrm{QP}_5$ & 1.015 & 31.414 & 0.9316 & 0.0960
&
& $\mathrm{QP}_5$ & 0.988 & 32.159 & 0.9397 & 0.0859 \\
\midrule

\multirow{5}{*}{\texttt{cut\_roasted\_beef}}
& $\mathrm{QP}_1$ & 3.974 & 34.538 & 0.9579 & 0.0457
&
\multirow{5}{*}{\texttt{sear\_steak}}
& $\mathrm{QP}_1$ & 3.197 & 34.898 & 0.9632 & 0.0377 \\
& $\mathrm{QP}_2$ & 3.073 & 34.454 & 0.9571 & 0.0481
&
& $\mathrm{QP}_2$ & 3.043 & 34.997 & 0.9625 & 0.0367 \\
& $\mathrm{QP}_3$ & 2.088 & 34.115 & 0.9543 & 0.0530
&
& $\mathrm{QP}_3$ & 1.933 & 34.637 & 0.9603 & 0.0412 \\
& $\mathrm{QP}_4$ & 1.344 & 33.343 & 0.9472 & 0.0676
&
& $\mathrm{QP}_4$ & 1.250 & 33.936 & 0.9547 & 0.0520 \\
& $\mathrm{QP}_5$ & 1.013 & 31.808 & 0.9325 & 0.0969
&
& $\mathrm{QP}_5$ & 0.974 & 32.475 & 0.9415 & 0.0817 \\

\bottomrule
\end{tabular}
\end{minipage}

\vspace{0.9em}

\begin{minipage}{\linewidth}
\centering
\textbf{(b) MPEG MIV}
\vspace{0.35em}

\begin{tabular}{rccccc@{\hspace{1.0em}}rccccc}
\toprule
\textbf{Sequence} & \textbf{QP} & \textbf{Size (MB)} & \textbf{PSNR} & \textbf{SSIM} & \textbf{LPIPS}
&
\textbf{Sequence} & \textbf{QP} & \textbf{Size (MB)} & \textbf{PSNR} & \textbf{SSIM} & \textbf{LPIPS} \\
\midrule

\multirow{5}{*}{\texttt{Painter (D01)}}
& $\mathrm{QP}_1$ & 3.151 & 36.093 & 0.9621 & 0.0307
&
\multirow{5}{*}{\texttt{Frog (E01)}}
& $\mathrm{QP}_1$ & 4.481 & 28.254 & 0.9006 & 0.1067 \\
& $\mathrm{QP}_2$ & 2.713 & 36.119 & 0.9621 & 0.0305
&
& $\mathrm{QP}_2$ & 3.746 & 28.217 & 0.8991 & 0.1102 \\
& $\mathrm{QP}_3$ & 1.896 & 35.477 & 0.9578 & 0.0340
&
& $\mathrm{QP}_3$ & 2.769 & 27.843 & 0.8927 & 0.1153 \\
& $\mathrm{QP}_4$ & 1.287 & 34.272 & 0.9458 & 0.0445
&
& $\mathrm{QP}_4$ & 1.748 & 26.700 & 0.8711 & 0.1386 \\
& $\mathrm{QP}_5$ & 0.959 & 32.281 & 0.9178 & 0.0719
&
& $\mathrm{QP}_5$ & 1.243 & 24.090 & 0.7953 & 0.2177 \\
\midrule

\multirow{5}{*}{\texttt{Carpark (E02)}}
& $\mathrm{QP}_1$ & 3.185 & 33.571 & 0.9332 & 0.0455
&
& & & & & \\
& $\mathrm{QP}_2$ & 2.601 & 33.445 & 0.9310 & 0.0475
&
& & & & & \\
& $\mathrm{QP}_3$ & 1.915 & 32.995 & 0.9252 & 0.0531
&
& & & & & \\
& $\mathrm{QP}_4$ & 1.326 & 31.814 & 0.9053 & 0.0732
&
& & & & & \\
& $\mathrm{QP}_5$ & 1.013 & 29.800 & 0.8583 & 0.1211
&
& & & & & \\

\bottomrule
\end{tabular}
\end{minipage}

\end{table*}

\subsection{Results on Challenging Sequences}
\label{app:hard}
Table~\ref{tab:hard} reports the uncompressed representations (the setting of Table~1) on the most challenging sequences
of each dataset: \textit{coffee\_martini} and \textit{flame\_salmon} from Neu3D, which contain specular surfaces, fast hand
motion and thin flames; Frog (E01) from MPEG MIV, with large non-rigid motion close to the cameras; and both Panoptic Sports
sequences, with fast full-body motion. On Panoptic Sports and E01, SAGA outperforms 4D Scaffold-GS by 1.6--2.4\,dB in PSNR,
and on Panoptic Sports even the small configuration (8.71\,MB) is 1.7--1.9\,dB better at 13--31$\times$ smaller size. On the
two Neu3D sequences, the large configuration is within 0.04\,dB (\textit{coffee\_martini}) and 0.34\,dB
(\textit{flame\_salmon}) of 4D Scaffold-GS in PSNR while being 5--8$\times$ smaller.

\begin{table}[h]
\centering
\caption{\textbf{Challenging sequences.} Uncompressed representations; sizes are raw (fp32) model sizes. The size of SAGA is
fixed by its anchor budgets and is the same for every sequence. Best PSNR and SSIM per sequence in bold.}
\label{tab:hard}
\resizebox{\linewidth}{!}{%
\begin{tabular}{llccccccccc}
\toprule
 & & \multicolumn{3}{c}{4D Scaffold-GS~\cite{cho20264d}} & \multicolumn{2}{c}{SAGA-S (8.71\,MB)} & \multicolumn{2}{c}{SAGA-M (17.37\,MB)} & \multicolumn{2}{c}{SAGA-L (34.54\,MB)} \\
\cmidrule(lr){3-5}\cmidrule(lr){6-7}\cmidrule(lr){8-9}\cmidrule(lr){10-11}
Dataset & Sequence & Size (MB)$\downarrow$ & PSNR$\uparrow$ & SSIM$\uparrow$ & PSNR$\uparrow$ & SSIM$\uparrow$ & PSNR$\uparrow$ & SSIM$\uparrow$ & PSNR$\uparrow$ & SSIM$\uparrow$ \\
\midrule
\multirow{2}{*}{Neu3D}
 & \textit{coffee\_martini} & 184.8 & \textbf{28.65} & \textbf{0.919} & 27.27 & 0.893 & 28.34 & 0.912 & 28.61 & 0.918 \\
 & \textit{flame\_salmon}   & 263.3 & \textbf{28.95} & \textbf{0.926} & 28.03 & 0.911 & 28.37 & 0.915 & 28.61 & 0.922 \\
\midrule
MPEG MIV
 & Frog (E01)               & 310.8 & 28.72 & 0.923 & 27.43 & 0.883 & 28.46 & 0.903 & \textbf{30.33} & \textbf{0.935} \\
\midrule
\multirow{2}{*}{Panoptic Sports}
 & \textit{basketball}      & 269.9 & 27.99 & 0.922 & 29.72 & 0.930 & \textbf{29.94} & \textbf{0.932} & 29.52 & 0.931 \\
 & \textit{boxes}           & 117.4 & 27.89 & 0.928 & 29.75 & 0.937 & \textbf{30.31} & \textbf{0.940} & 28.82 & 0.935 \\
\bottomrule
\end{tabular}}
\end{table}

\section{Additional Qualitative Results}

Here we provide exemplary apple-to-apple comparisons between the source frames and reconstructions. More visual comparisons with different camera motions are available on the \href{https://ge1gao.github.io/saga/}{\textcolor{lilac}{\textbf{project page}}}.

\begin{figure*}[t]
  \centering

  \begin{minipage}{\linewidth}
    \centering
    \includegraphics[width=\linewidth]{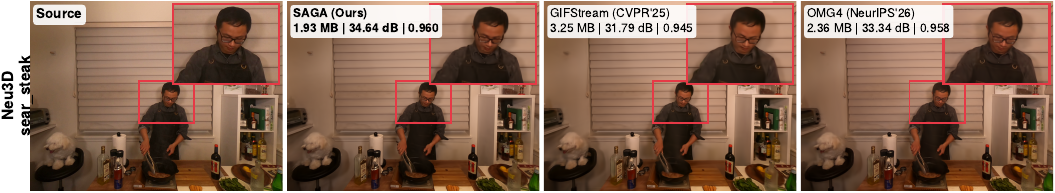}
    \caption{Qualitative comparison on Neu3D \textit{sear\_steak}, frame 150.}
    \label{fig:qual_extra_sear_steak}
  \end{minipage}

  \vspace{0.8em}

  \begin{minipage}{\linewidth}
    \centering
    \includegraphics[width=\linewidth]{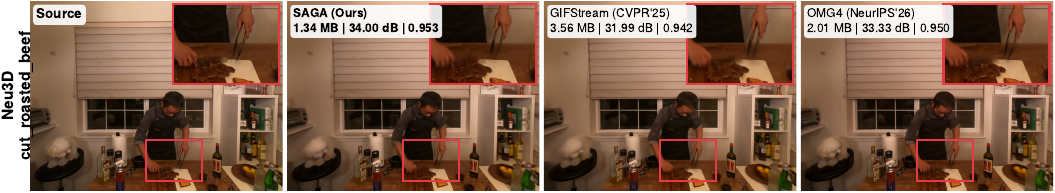}
    \caption{Qualitative comparison on Neu3D \textit{cut\_roasted\_beef}, frame 150.}
    \label{fig:qual_extra_cut_roasted_beef}
  \end{minipage}

  \vspace{0.8em}

  \begin{minipage}{\linewidth}
    \centering
    \includegraphics[width=\linewidth]{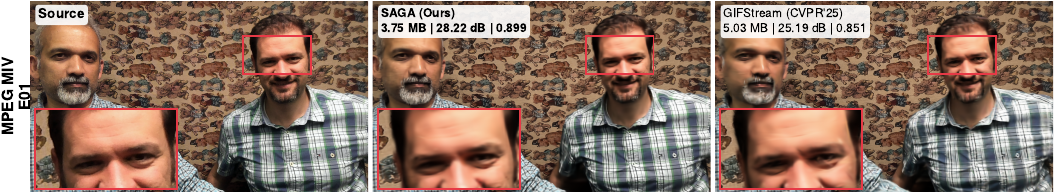}
    \caption{Qualitative comparison on MPEG MIV \textit{E01}, frame 32.}
    \label{fig:qual_extra_miv_e01}
  \end{minipage}

  \vspace{0.8em}

  \begin{minipage}{\linewidth}
    \centering
    \includegraphics[width=\linewidth]{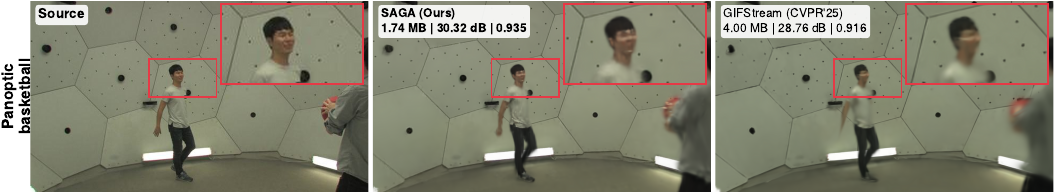}
    \caption{Qualitative comparison on Panoptic \textit{basketball}, frame 75.}
    \label{fig:qual_extra_basketball}
  \end{minipage}

  \vspace{0.8em}

  \begin{minipage}{\linewidth}
    \centering
    \includegraphics[width=\linewidth]{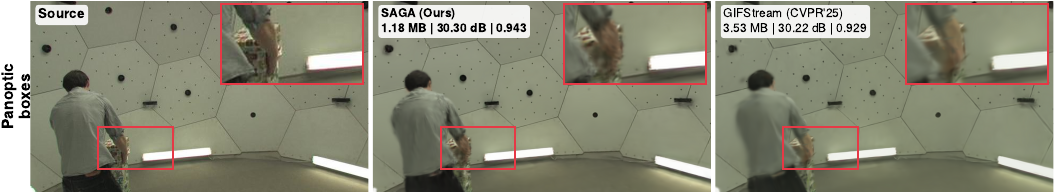}
    \caption{Qualitative comparison on Panoptic \textit{boxes}, frame 75.}
    \label{fig:qual_extra_boxes}
  \end{minipage}

\end{figure*}


\end{document}